\documentclass[twocolumn]{article}
 \usepackage{amssymb}
 \usepackage{amsmath}
 \usepackage[colorlinks=true, citecolor={blue}]{hyperref}
 \usepackage{graphicx}
\usepackage{float}
\usepackage{booktabs}
\usepackage{multirow}
\usepackage{tabularx}
\usepackage{adjustbox}
\usepackage{booktabs}

\usepackage{array}
\newcolumntype{$}{>{\global\let\currentrowstyle\relax}}
\newcolumntype{^}{>{\currentrowstyle}}
\newcommand{\rowstyle}[1]{\gdef\currentrowstyle{#1}%
    #1\ignorespaces
}

\begin{document}

\title{A machine learning approach for underwater gas leakage detection}
\author{Paulo Hubert$^1$ \\
Linilson R. Padovese$^1$ \footnote{The authors gratefully acknowledge University of S\~{a}o Paulo and support from SHELL Brazil (subsidiary company of Royal Dutch Shell) and FAPESP, through the Research Centre for Gas Innovation (RCGI) hosted by the University of S\~{a}o Paulo (FAPESP Grant Proc. 2014/50279-4). We would also like to thank FAPESP and CNPq for their support, by grants number FAPESP 2016/02175-0  and CNPq 303992/2017-4.}}
\date{%
    $^1$Department of Mechanical Engineering, Escola Polit\'{e}cnica - University of S\~{a}o Paulo, S\~{a}o Paulo, SP - Brazil\\[2ex]%
}

\maketitle



\abstract{
Underwater gas reservoirs are used in many situations. In particular, Carbon Capture and Storage (CCS) facilities that are currently being developed intend to store greenhouse gases inside geological formations in the deep sea. In these formations, however, the gas might percolate, leaking back to the water and eventually to the atmosphere. The early detection of such leaks is therefore tantamount to any underwater CCS project. In this work, we propose to use Passive Acoustic Monitoring (PAM) and a machine learning approach to design efficient detectors that can signal the presence of a leakage. We use data obtained from simulation experiments off the Brazilian shore, and show that the detection based on classification algorithms achieve good performance. We also propose a smoothing strategy based on Hidden Markov Models in order to incorporate previous knowledge about the probabilities of leakage occurrences.}

\emph{keywords:} CCS ; Underwater acoustics ; Signal detection ; Machine Learning

\section{Introduction} \label{sec:intro}

From the past recent years, CO2 capture and storage (CCS) technology has been considered to be a game-changing technology to avoid human-induced global warming and the resulting climatic change\cite{EU2009}.

There are many challenges that must be met in order to guarantee the safety of the geologic reservoirs used to store greenhouse gases. One of them is to avoid and monitor leakages from the reservoir.

International literature describes plenty of monitoring tools that have been tested and have been used in the last years for marine CO2 storage monitoring programs \cite{IEAGHG2015, Fasham2015}. Some of them are used for rapid and focused spatial monitoring; others are intended for long time and large area coverage. 

Regarding passive acoustic monitoring, when these leakages arise in the form of bubbles a characteristic acoustic signal is produced, as shown by \cite{Minnaert1933, Strasberg1956, Berges2015, Miao2018}. This signal can be used for detecting and locating gaseous leaks.

This work proposes the development of a Passive Acoustic Monitoring (PAM) system for leakage detection on offshore CO2 geological storages and facilities. The characteristic signal caused by the acoustical emission of bubbles is explored in the design of classification models, by using simulated leakages obtained experimentally. The main advantages of using PAM in signal detection are the relatively low cost of the sensoring equipment, and the long range of the sensor, especially when detecting low-frequency acoustic signals.

This paper is organized as follows: section \ref{sec:class} discusses the use of classification algorithms for signal detection; section \ref{sec:exp} describes the pilot experiment conducted to obtain data from simulated leakages. Section \ref{sec:train} describes the training of the classifiers, and section \ref{sec:hmm} proposes a smoothing procedure using Hidden Markov Models that uses the classifier predicitons to obtain a detection system. Section \ref{sec:conc} concludes the paper.

\section{Classification algorithms for signal detection} \label{sec:class}

Traditional signal detection procedures are based on a thorough analysis of the phenomenon of interest, and the signal it induces on the sensor. In the case of leakage detection and PAM, this means analysing the acoustic emission model of gas bubbles in water \cite{Berges2014}. After analyzing the signal, the detector works usually by imposing some statistical model on the background noise field, and then designing an efficient estimator or statistical hypothesis testing procedure to obtain, given a sample from the sensor, the probability that the signal of interest is actually present in the data.

This approach has the advantages of working from first principles (i.e., from a physical model describing the phenomenon), and also of having (in principle) no need of experimental data, particularly of background noise examples. If an accurate physical model is available, and if the probabilistic model for the background noise is general enough, it is possible to design good detectors using this approach \cite{Davis1989}.

There are some drawbacks, however, in the classical detection approach. First, there is the complexity of the physical model, which in many cases can be very challenging to solve, or even unsolvable.  Also, when deployed in real operational conditions, these detectors suffer from a large computational burden: since the actual instant of the beginning of the signal is unknown (i.e., the signal's phase is unknown), the detector must be applied to the entire sensor data, usually using sliding windows or some similar method. This can make the detection procedure very cumbersome; see for instance a past work from the authors \cite{Hubert2017}, where a Bayesian testing procedure is applied to boat detection on underwater acoustic data. 

The use of classification algorithms overcomes this drawback, in exchange of demanding previously annotated samples. When using this approach, the physical model can be ignored; furthermore, even if the training step of the algorithm is computationally intensive, the actual application of the detection algorithm is fast, usually depending only on a forward pass of a fixed length sample signal through the pre-trained algorithm.

There is however a critical question in the use of classification algorithms for signal detection: the availability of negative examples, which include samples of background noise only but also (ideally) samples from different events that might confound the detector. It is in principle possible that the algorithm will learn to distinguish the noise from the signal by acquiring a precise representation of the noise, instead of representing the signal. In other words, a classification algorithm might learn features of the noise and use them to correctly classify the signal, leading to detectors with very high accuracy on the training set, but with little generalization power, especially if the noise samples are not representative enough of the full range of operational conditions of the sensor.

The best way to avoid this problem is to spend time and effort in building a large and rich set of samples from different conditions and with the presence of many different events. In some situations, where the operational conditions of the detector are well-known and reasonably well-behaved, it might be practical to build such a set of training examples. This is the case, for instance, in the problem of leakage detection in deep ocean waters, where not many confounding events are expected and where the background noise field conditions are reasonably stable. When this database of negative examples is not available, the tuning of the algorithm, and the actual use of the algorithm's prediction in the detector's design must be made with extreme care.

In this work, we train our classification-detector algorithm on samples of background noise and background noise plus leakage only. To avoid fitting the detector to a specific background noise field, we will separate our noise samples in training and test sets based on the time of the day when the recordings were taken, in order to guarantee that the algorithm is tested against different background conditions. We acknowledge, however, that our data is not representative of the full variability of acoustical events in subaquatic environments; we intend to investigate this question in more depth after further experiments are conducted.

The algorithms that will be used to build the classifier-detector are the Random Forests algorithm and the Gradient Boosting Trees algorithm. Random Forests have been previously applied to acoustic events detection \cite{Phan2015}, specifically in the context of speech recognition. Gradient Boosted Tress have also been applied to acoustic signal analysis \cite{Fonseca2017}, but to the best of our knowledge not to the design of detectors for specific events.

\subsection{The Random Forests algorithm for classification}

The CART (Classification and Regression Trees) algorithm was first proposed by Breiman \emph{et al} \cite{Breiman1984}. It is based on the simple idea of recursively partitioning the feature space in a set of rectangular regions, where each new partition is based on the value of a single variable. The classification is done by applying a majority vote rule to each region obtained after the partition is finished.

Even though the CART algorithm was efficient in solving many classification tasks, the fitting (or training) algorithm was sensible to small changes in the dataset (what in the machine learning and statistics community is called \emph{high variance} of the classifier's predictions). 

To control the variance of the algorithm, ensemble methods have been proposed. Ensemble methods try to improve the performance of a given class of algorithms by training multiple instances of the algorithm on subsets of the data, and then combining the resulting predictions of each weaker model.

Tin Kam Ho \cite{Ho1995} first proposed an ensemble method based on classification trees; he uses a random subspace approach, where different trees are trained on a random subset of the available features. Later, Breiman \cite{Breiman2001} extended Ho's method by also including a bootstrap aggregation step, where individual samples from the training set are also randomly selected to be used in each model. The resulting algorithm was called \emph{Random Forests} by Breiman. 

There are now many available implementations of the Random Forests algorithm. In this work, we use the python implementation from the \emph{scikit-learn} toolset available in \url{https://scikit-learn.org}.

\subsection{Gradient Boosted Trees}

The ideia of \emph{boosting} a learning algorithm developed from the investigation of the possibility of combining weak learning algorithms to form a strong learner \cite{Schapire1990}. The first boosting algorithm, \emph{AdaBoost}, was developed in 1997 by Freund and Schapire \cite{Freund1997}.

Gradient Boosted Machines were later developed by Jerome Freidman, among others, who noticed that boosting can be seen as a gradient descent procedure in a functional space \cite{Mason1999}. 

The main difference between Random Forests and Gradient Boosted Trees is that in Random Forests several weaker models are trained in parallel (i.e., each model is trained without regard for the results of every other model), whilst in Gradient Boosting the models are trained in a sequential manner, each model feeding on the last one's results. 

We adopt the \emph{XGBoost} implementation of Gradient Boosted Trees (\url{http://https://github.com/dmlc/xgboost}), which is available for many computing platforms, including python and R.

\section{Experimental setup}\label{sec:exp}

An experimental sea campaign was planned and carried out in order to obtain a set of experimental data to validate the leak detection algorithms. The leakage was simulated through the use of compressed air (from scuba dive cylinders), with flow, pressure and exit diameter orifice controlled. These controlled leaks were performed at predetermined distances from underwater acoustics monitoring equipment

In this first experimental campaign, the difference in pressure between the cylinder and tube outlet was kept constant at $9$ bar. The flow rates used were three: $2$, $5$ and $10$ l / min. The distances between the leakage nozzle and hydrophones was $3\, m$.

The underwater acoustic monitoring system consists of one hydrophone developed by the laboratory, with a flat frequency between $5$ Hz and $50$ kHz, and sensitivity of $-154$ $dB$ $rel$ $1$ $\mu Pa$. The digitization of the acoustic signals was carried out by a TASCAM-800 audio interface connected to a notebook, using a sampling rate of $48$ $kHz$. Both the leakage outlet and hydrophones were positioned at $8$ $m$ depth.

In this pilot experiment, the acquired data has a total duration of approximately $30 \, min$, obtained through a period of roughly $3$ hours in the sea.  

\section{Training and testing the classification algorithms}\label{sec:train}

Our experimental data contains examples of the simulated leakage with three different gas flux intensities ($2$, $5$ and $10\, l/min$). The dataset contains a total amount of $1,900$ seconds of recordings, where $1,555$ seconds were taken with the bubble generator turned on, and the remaining $345$ seconds were taken with the bubble generator turned off.

To evaluate the performance of the classification algorithms, we chose to train them using only the samples where the bubble flow was the largest. The rationale behind this choice is that this experimental condition is the best for training a detector, since these are the strongest signals in our dataset. Aditionally, we are interested in analyzing the performance of the algorithm trained on high signal-to-noise ratio (SNR) data, when applied to detection of leakages with smaller SNR, i.e., with a lower flow of gas. This reduces the total signal length (for training) to $476$ seconds. 

After separating the signal's examples to train the algorithm, we must also choose a set of negative examples, i.e., examples of background noise. This is a critical choice, as discussed above; we would like to be able to verify if the classifier is not taking advantage of a precise representation of the noise samples. 

In our dataset there are a few recordings taken at different times of the day. We admit that, during a given continuous recording, the background noise characteristics will be more homogeneous than between different recordings taken at different times. Therefore we adopt the following strategy: to train the classifier, we use a set of negative examples taken from the same continuous recordings, and to test it we use a different set, recorded later on the same day. Doing this we guarantee that our sets of negative examples are maximally different in the training and testing samples. 

After this separation, our full training dataset contains $742\,s$ of signal, where $476\, s$ contain the signal and $266 \,s$ are noise-only examples.

The training signal is further divided into smaller sections that will be used as the actual sample units in the classifier design. We test windows with different sizes and with different overlap values.

For each window size, we train the classifier using as features a) the signal's periodogram, and b) the power spectral density (PSD) smoothed estimate using Welch's method with Hann windows. We filter both the periodogram and PSD to the band $150-500$ $Hz$, which is the band where the leakage acoustic emission is expected to be found.

\subsection{Selection of the classification algorithm}\label{sec:selection}

To train the Random Forests (RF) and Gradient Boosted Trees (GBT) we start by running grid searches to obtain best values for the hyperparameters of each algorithm. The grid search is based on a $5$-fold  cross-validation on the training set. The results are shown in table \ref{tbl:res1}.

\begin{table*}[h]
\centering
\caption{Selection of classifier and feature; see text for details}
\label{tbl:res1}
\begin{tabular}{$r^r^r^r^r}
\toprule
 Duration (s) &  Overlap (s) & Algorithm & Feature & Accuracy CV \\
\midrule
            1 &          0.0 &       xgb &   welch &       0.747 \\
            1 &          0.5 &        rf &   welch &       0.729 \\
            2 &          0.0 &       xgb &   welch &       0.781 \\
            2 &          1.0 &        rf &   welch &       0.767 \\
            3 &          1.0 &        rf &   welch &       0.787 \\
            3 &          2.0 &        rf &   welch &       0.787 \\
            4 &          2.0 &        rf &   welch &       0.813 \\
\rowstyle{\bfseries}             4 &          3.0 &       xgb &   welch &       0.817 \\
            5 &          3.0 &        rf &   welch &       0.802 \\
            5 &          4.0 &       xgb &   welch &       0.808 \\
\bottomrule
\end{tabular}
\end{table*}

The best cross-validation performance was shown by the Gradient Boosted Trees algorithm, working with Welch estimates of the PSD on $4$ seconds windows with $3$ seconds overlap. The Random Forests algorithm working on $4$ seconds windows with $2$ seconds overlap had practically equivalent results.

As a general rule, algorithms trained on longer windows show better accuracy, and the Random Forests algorithm performs better in $6$ out of the $10$ investigated scenarios. Also, the use of Welch estimates of the PSD provides better results than using the periodogram in all cases.

To further analyze the performance of the tested algorithms, we apply them to the classification of samples from different flows ($2$ and $5$ l/minute). The results are shown in table \ref{tbl:res2}.
 
\begin{table}
\centering
\caption{Precision of the classifiers applied to different leakage flows}
\label{tbl:res2}
\begin{tabular}{$r^r^r^r}
\toprule
Algorithm & Feature &  Flow &  Precision \\
\midrule
\rowstyle{\bfseries}     rf &     psd &     2 &   0.862 \\
\rowstyle{\bfseries}     rf &     psd &     5 &   0.979 \\
   rf &   welch &     2 &   0.702 \\
   rf &   welch &     5 &   0.901 \\
  xgb &     psd &     2 &   0.665 \\
  xgb &     psd &     5 &   0.896 \\
  xgb &   welch &     2 &   0.713 \\
  xgb &   welch &     5 &   0.872 \\
\bottomrule
\end{tabular}
\end{table}

As expected, the precision was always higher on the samples with greater flows ($5$ l/min). But in both cases of different flows, the best algorithm was the Random Forests, using the periodogram estimator of the PSD.

These results indicate that the Random Forests classifier generalises better than the GBT for different flows. This fact deserver a deeper analysis, which we intend to present in a future work where we will investigate the use of machine learning algorithms to quantify (not only detect) the leakages.

As for the detection performance, both classification algorithms show promising results, achieving a good precision in cross-validation and also when applied to different flow rates. 

For the remainder of the paper we pick the GBT algorithm using Welch as the classifier of choice; considering the present goal (detection), we consider the cross-validation results as more important than the test using different flow rates.

The next step after selecting the best algorithm for the classification of individual signal windows is to actually use its predictions to build a detector. This will be discussed in the next section.

\section{Detector design: classification and HMM smoothing}\label{sec:hmm}

The classification algorithm applied to a new signal produces a prediction score in $[0,1]$, where higher values can be interpreted as higher evidence for the presence of a leakage in the given signal. Usually, a threshold is applied: when the score is higher than a given constant (often $0.5$), the signal is classified as leakage; otherwise, it is classified as noise only.

Choosing a higher threshold to classify a given window as a leakage has the immediate effect of decreasing the false alarm ratio, but at the expense of also decreasing the true positive ratio. On the other hand, choosing a low threshold has the opposite effect. The choice of threshold, then, must consider the balance between the two goals: minimize false alarms while also maximizing detection probability.

By applying the threshold to the training data, it is possible to estimate (via confusion matrix) the accuracy measures of the resulting classifier. In the next section, we propose to use these estimates in a  Hidden Markov model to smooth the algorithms predictions while at the same time incorporating domain-based knowledge about the occurrence of leakages.

\subsection{Hidden Markov model for the occurrence of leakages}

Consider that the presence or absence of a leakage is a hidden binary variable which we want to infer. Call this variable $X_t$, where $X_t = 1$ if there is a leakage at time $t$, and $X_t = 0$ otherwise.

Suppose that  $X_0 = 0$. Then, at any instant, a leakage can start; in this case, the stochastic process $X_t$ suffers a transition from state $0$ to state $1$.  If there is no leakage starting between $t$ and $t+1$, the process stays in the same state (i.e., there is a transition from $X_t = 0$ to $X_{t+1}= 0$).

Likewise, whenever a leakage is occurring ($X_t = 1$), there is the possibility that it spontaneously stops ($X_{t+1} = 0$); if it doesn't, the leakage continues (i.e., $X_{t+1} = 1$).

We propose to model this process as a (hidden) Markov chain, with transition probabilities given by $P(X_{t+1} = 1 | X_t = 0) = \epsilon$ and $P(X_{t+1} = 0 | X_t = 1) = \delta$. In this model, $\epsilon$ represents the probability of a leakage starting at a given time $t$, and $\delta$ represents the probability that a leakage is spontaneously repaired. This possibility is rather unlikely, and this can be induced in the model by adopting a small value for $\delta$. 

The hidden Markov model (HMM) can be completed in the following way: at any given time $t$, the classification is applied to the signal, yielding a prediction $Y_t$. $Y_t$ can be either $1$ (leakage detected) or $0$ (no detection). By observing the cross validation results from the classifiers, we can estimate the corresponding \textbf{emission} probabilities $P(Y_t = 1 | X_t = 1)$ as the positive recall, and $P(Y_t = 0 | X_t = 0)$ as the negative recall of the classification algorithm. These probabilities depend upon the threshold used to generate class predictions.

After fully defining the HMM, it is possible to calculate the probability that a leakage is actually occurring, given a sequence of predictions from the classifier, that is, $P(X_t = 1 | Y_1, Y_2, ..., Y_t)$. Defining $\pi_t(x) = P(X_t = x | Y_1, Y_2, ..., Y_t)$, this can be accomplished by the usual forward recursion formula:

\begin{equation}
\pi_t(X_t) = P(Y_t | X_t)\sum_{X_{t-1}}P(X_t | X_{t-1})\pi_{t-1}(X_{t-1})
\end{equation}

\subsection{Testing the detector}

To test the full detection strategy, we take the following steps:

\begin{enumerate}
	\item Using a training-test sample split, select the best classification algorithm by cross-validation;
	\item The best classification algorithm is trained on a subset of the dataset, excluding a continuous section of our signal to be used in the detector test;
	\item The classifier is applied to the test sample and predicted probability values are obtained;
	\item A threshold value is chosen to turn the probability predictions into class predictions. The same threshold is used to estimate the positive and negative recall of the algorithm using the training set;
	\item The class prediction values are smoothed by the forward recursion algorithm, yielding the detector values.
\end{enumerate}

\begin{figure*}[ht]
 \centering
 \includegraphics[width=1.\textwidth]{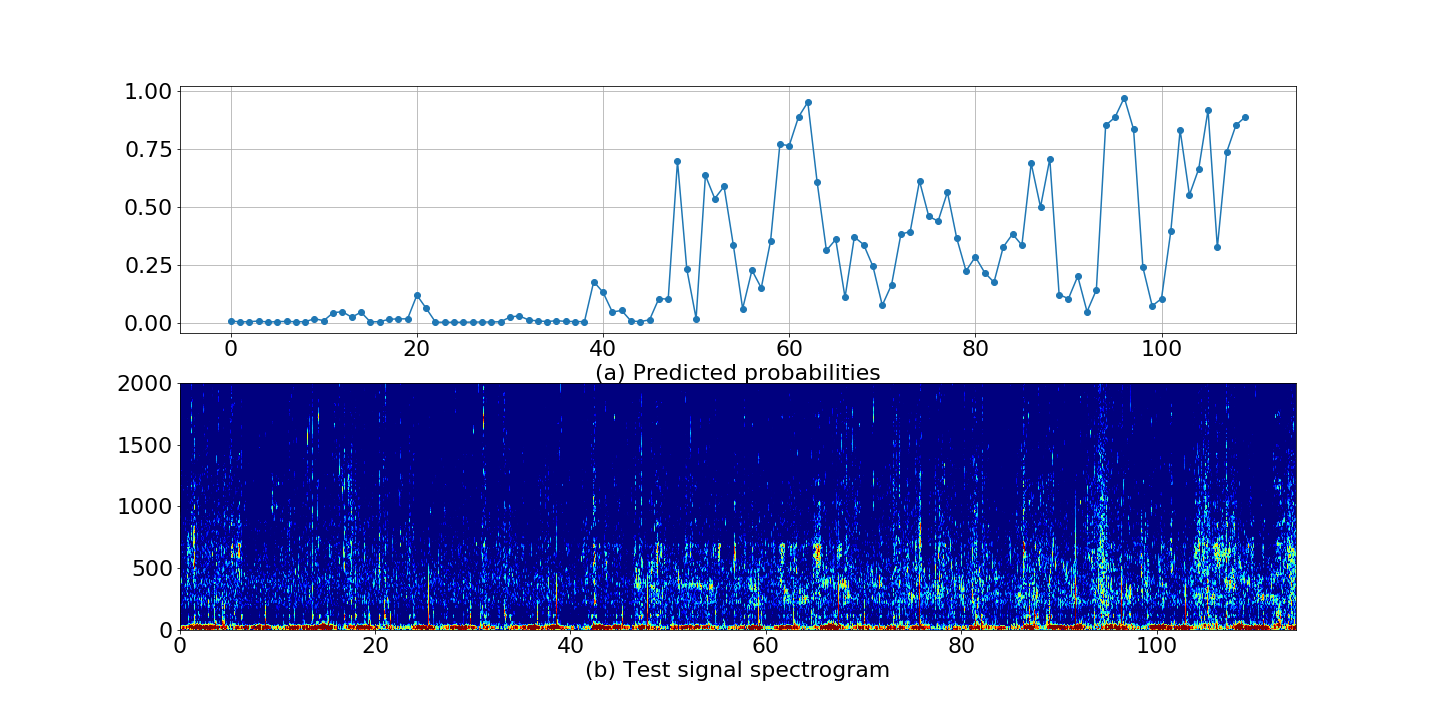}
 \caption{Predicted probabilites and spectrogram of test signal}
 \label{fig:detect_prob}
\end{figure*}

Item \emph{b} of figure \ref{fig:detect_prob} shows the spectrogram of the test signal selected to test the classification-detection approach. The simulated leakage starts at $\sim 48$ seconds. 

The first step (algorithm selection) has been done and the results reported on section \ref{sec:selection}. For the second step, we train the selected algorithm using all available data except the section of the signal to be used as test data. Applying the trained model to the test signal yields the predicted probabilities shown in figure \ref{fig:detect_prob}, item $a$.

%

\begin{figure*}[ht]
 \centering
 \includegraphics[width=1.\textwidth]{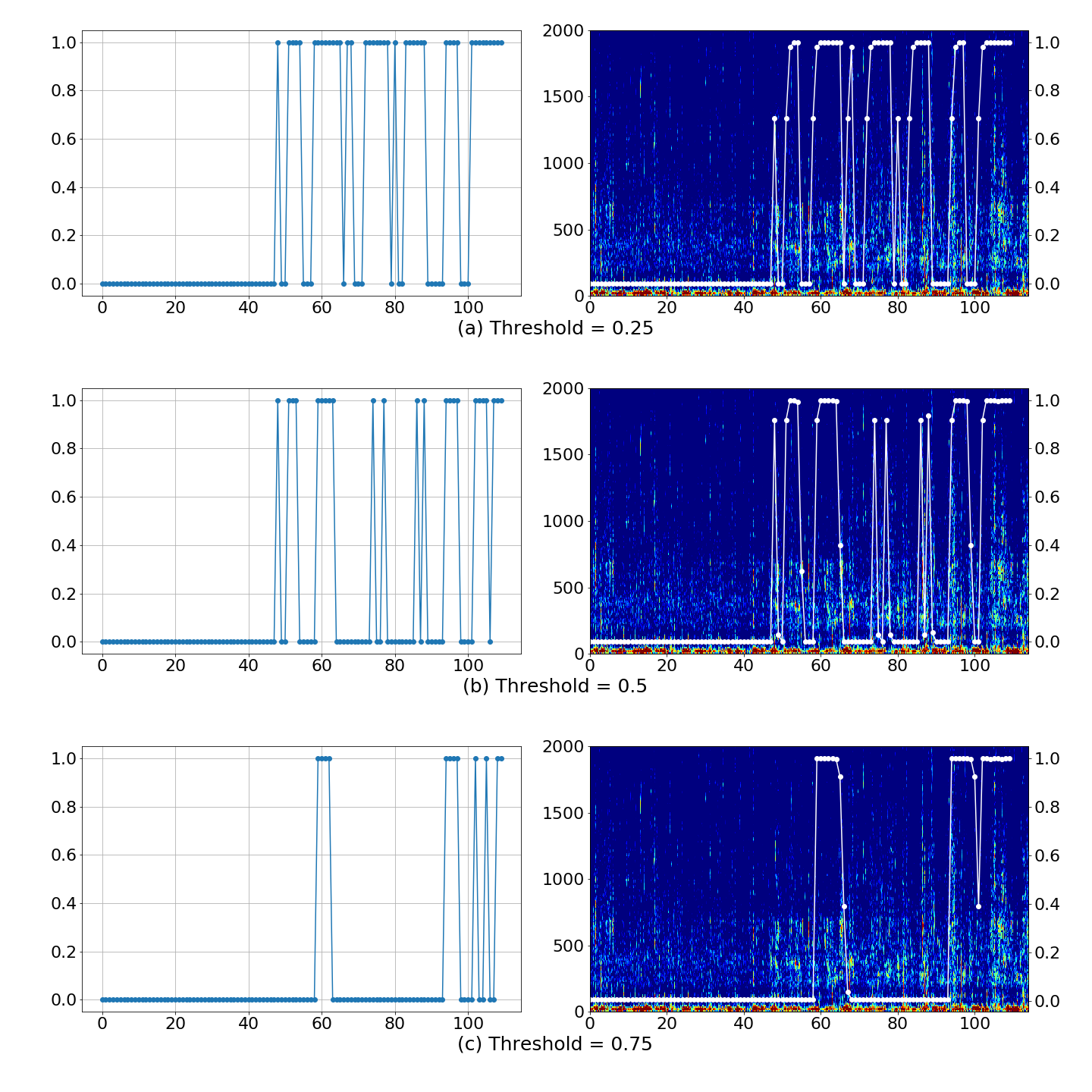}
 \caption{Predicted class and HMM probabilities}
 \label{fig:detect_final}
\end{figure*}

Next, to obtain the emission probabilities for the HMM, we first choose a threshold for the predicted probability and then apply a $10$-fold cross validation of our selected model on the training set.  With the cross validation results we are able to estimate both the positive recall (probability of detection) and negative recall (the reciprocal of the probability of false alarm). Figure \ref{fig:detect_final} shows the class predictions (obtained by the application of the threshold over the predicted probabilities) for the test signal and the resulting HMM probabilities of leakage, for $3$ values of the threshold. For the transition probabilities, we adopt $\delta = 0.00001$ and $\epsilon = 0.1$. The value of $\delta$ is chosen to reflect the fact that it is highly unlikely that a true leakage will stop spontaneously.

The effect of applying the HMM over the class predictions depends on the estimates of the positive and negative recall (and thus depends on the choice of threshold). With the lowest threshold of $0.25$, the HMM smoothing causes the detector to delay response to a positive identification from the classifier: the HMM smoothed values reach $0.7$ first, and only after two consecutive positive identifications the probability of a leakage reaches $1.0$. On the other hand, a single negative result from the classifier causes the detection probability to immediately drop to $0$. The main cause for this behavior is the high value of the probability of detection: since $P(Y_t = 1 | X_t = 1)$ is very high, the reciprocal $P(Y_t = 0 | X_t = 1)$ is close to $0$; so when confronted with a $0$ prediction from the classifier, the HMM admits that it must be a true negative and drops $P(X_t = 1 | Y)$ accordingly.

When the threshold is raised to $0.5$, the effect of a negative prediction from the classifier is also delayed: it does not lead immediately to a $0$ probability value in the detector. This is mainly due to the decrease in the value of the positive recall (probability of detection): the HMM is now less confident that, if a leakage is happening, the classifier would have detected it. Thus, if it sees a $0$ prediction by the classifier following a $1$, it admits that the $0$ might be a false negative (which is more likely, now that the probability of detection has dropped).

Finally, when the threshold is the highest ($0.75$), the effect of negative predictions from the classifier ends up being completely smoothed out after a few positive predictions. The detector probability will only drop if many negative predictions appear sequentially. This can be seen at the final end of the signal.

The choice of the final threshold to be implemented in the detector system must take into account the relative costs of issuing a false alarm, and letting a leakage remain undetected. Given the fact that a true leakage is a long duration acoustic event (which is reflected on the small probability of a $1-0$ transition in the hidden Markov chain), it might be advisable to pick high values for the threshold. This will decrease the false alarm ratio, and, if the classifier is efficient, will still correctly capture true leakages, because in this case the positive predictions will accumulate over time and the HMM will also accumulate the evidence, yielding a consistently high probability.

\section{Conclusion}\label{sec:conc}

Our main goal in this paper was to investigate the viability of applying machine learning algorithms to the task of underwater gas leakage detection. We analyzed the performance of two algorithms, Random Forests and Gradient Boosted Trees, using data from a pilot study with simulated leakages. We have also proposed to use a hidden Markov model to incorporate knowledge about the duration of actual leakages, in particular the fact that once a leakage takes place, there is a very small probability that it will spontaneously stop. 

The results show that this strategy is promising. The final classifier algorithm showed good performance, even though it was trained in a relatively small sample. Also, the use of the hidden Markov model allows the detector to incorporate knowledge about the occurrence and duration of leakages, and also incorporates knowledge about the classifier's performance (the positive and negative feedback rates).

For future works we intend to investigate other classification strategies. Other lines of work involves the study of more precise methods to estimate the PSD of a given signal, and the analysis of complete probabilistic models that combine the classifier and HMM smoother in a single model.

We are also conducting new experiments to enrich our data set. The new data will support both the training of more powerful classifiers and the investigation of yet another methods for leakage detection and quantification using machine learning algorithms.

\bibliographystyle{plain}
\bibliography{bibliografia}

\end{document}